\documentclass[runningheads]{llncs}
\usepackage{graphicx}
\usepackage{multirow}  
\usepackage{amsfonts} 
\usepackage{amsmath} 
\begin{document}
\title{LAMP: Large Deep Nets with Automated Model Parallelism for Image Segmentation}
\titlerunning{LAMP: Large Deep Nets with Automated Model Parallelism}
%
\author{Wentao Zhu \and Can Zhao \and Wenqi Li \and Holger Roth \and Ziyue Xu \and Daguang Xu}
\authorrunning{Wentao Zhu et al.}
%
\institute{NVIDIA} 
\maketitle              
\begin{abstract}
Deep Learning (DL) models are becoming larger, because the increase in model size might offer significant accuracy gain. To enable the training of large deep networks, data parallelism and model parallelism are two well-known approaches for parallel training. However, data parallelism does not help reduce memory footprint per device. In this work, we introduce Large deep 3D ConvNets with Automated Model Parallelism (LAMP) and investigate the impact of both input's and deep 3D ConvNets' size on segmentation accuracy. Through automated model parallelism, it is feasible to train large deep 3D ConvNets with a large input patch, even the whole image. Extensive experiments demonstrate that, facilitated by the automated model parallelism, the segmentation accuracy can be improved through increasing model size and input context size, and large input yields significant inference speedup compared with sliding window of small patches in the inference. Code is available\footnote{https://monai.io/research/lamp-automated-model-parallelism}.

\end{abstract}
\section{Introduction}
Currently, deep learning models have been becoming larger. More and more studies demonstrate that, the increase in model size offers significant accuracy gain. 
In the natural language processing (NLP), transformers have paved the way for large models. For instance, the Bert-large model~\cite{devlin2018bert} consumes 0.3 billion (B) parameters and GPT-2~\cite{radford2019language} has 1.5B parameters. In the image classification of computer vision, AmoebaNet (B)~\cite{huang2019gpipe} consists of 550 million (M) parameters and achieves the best top-1 accuracy of 84.4\% on ImageNet 2012 validation dataset~\cite{deng2009imagenet}. As the model size continues to grow, training these large models becomes challenging because it is difficult to fit the training within the memory limit of one single GPU.

There are several ways to train large models on GPUs. Model compression, such as mixed precision training~\cite{micikevicius2017mixed}, tries to use less bits to represent the network. It can reduce GPU memory consumption to some extent, however, might affect accuracy and can only fit a slightly or moderately large model to one GPU. Checkpointing~\cite{chen2016training,martens2012training} reduces the memory of the intermediate feature maps and gradients during training, such that the memory consumption can be reduced to $O(\log n)$ with $O(n \log n)$ extra time for forward computation in the network of $n$ layers theoretically. Invertible networks~\cite{gomez2017reversible,blumberg2018deeper,brugger2019partially,zhuang2019invertible} further reduce memory consumption to $O(1)$ by modifying the networks to be invertible which recalculate the feature maps in the back-propagation and might impact accuracy for discriminative models such as commonly used U-Net for segmentation~\cite{ronneberger2015u}.

Facilitated by the high speed communication tools such as NVLINK, parallel training across devices is a popular direction for this challenge. Generally, there are two common parallelisms to fit large models into GPUs without information loss and re-calculation, data parallelism and model parallelism~\cite{huang2019gpipe,rajbhandari2019zero,narayanan2019pipedream}. Data parallelism duplicates the model and runs split batch in multiple devices. It does not reduce model's memory footprint per device and cannot address out of memory issue faced by training large models. Model parallelism splits a model into multiple partitions and naturally handles this issue. For instance, a state-of-the-art model parallelism, Megatron, can scale up to 20B parameter models by using 16 GPUs. Advanced model parallelism executes partitions concurrently across devices for efficient training, and multiple model parallelisms have emerged, e.g., pipeline parallelism in GPipe~\cite{huang2019gpipe} and PipeDream~\cite{narayanan2019pipedream}, and TensorSlicing~\cite{shazeer2018mesh} in Megatron~\cite{shoeybi2019megatron} and Mesh Tensorflow~\cite{shazeer2018mesh}. However, model parallelisms, such as Megatron~\cite{shoeybi2019megatron}, only support a limited set of operators and models. 
For example, in medical image analysis, the most widely used model, U-Net~\cite{ronneberger2015u}, is not supported by these existing parallelisms.
In medical domain, it is a common need to be able to handle 3D volumetric image, which essentially consumes more memory with 3D ConvNets than their 2D counterparts. Unfortunately, current medical image computing is still limited by GPU memory size.  A lot of techniques, such as sliding window and resampling, are utilized to get around the problem. Moreover, the designed 3D models often use much less filters than advanced 2D models in each convolution~\cite{isensee2019nnu}. Therefore, insightful investigations of large models and large context, i.e., large input, might be extremely useful for the current research by leveraging automated model parallelism. 

Training large models with large input is especially challenging for medical images due to limited number of training data.
Large input increases context which is critical for image understanding~\cite{isensee2019nnu}. However, it reduces the variation of training input and aggravates the extremely imbalance issue among background and relatively small subjects (e.g., small organs and lesions) commonly existed in medical image computing~\cite{zhu2019anatomynet,sudre2017generalised}. Various loss functions have been proposed to alleviate this challenge. For example, adaptive weighted loss is proposed with a hybrid loss between dice loss of class-level loss and focal loss of voxel-level loss for small organ segmentation~\cite{zhu2019anatomynet}. 
The second example is the boundary loss~\cite{kervadec2019boundary}, which is different from previous approaches using unbalanced integrals over the regions. It uses integrals over the boundary (interface) between the regions, which can be implemented by a level set distance map weighted cross entropy loss leveraging an integral approach to computing boundary variations.
Transfer learning by fine-tuning from a pretrained model is another way to reduce the training difficulty of specially designed medical image models~\cite{tajbakhsh2016convolutional}. Based on learning theory such as curriculum learning~\cite{bengio2009curriculum,jesson2017cased}, a model can be well trained by firstly being fit easy samples/tasks and later being fit hard samples/tasks.

\subsubsection{Contributions}
\begin{figure}[t]
\begin{minipage}{0.5\linewidth}
		 			\includegraphics[width=\textwidth,trim=2 2 2 2,clip]{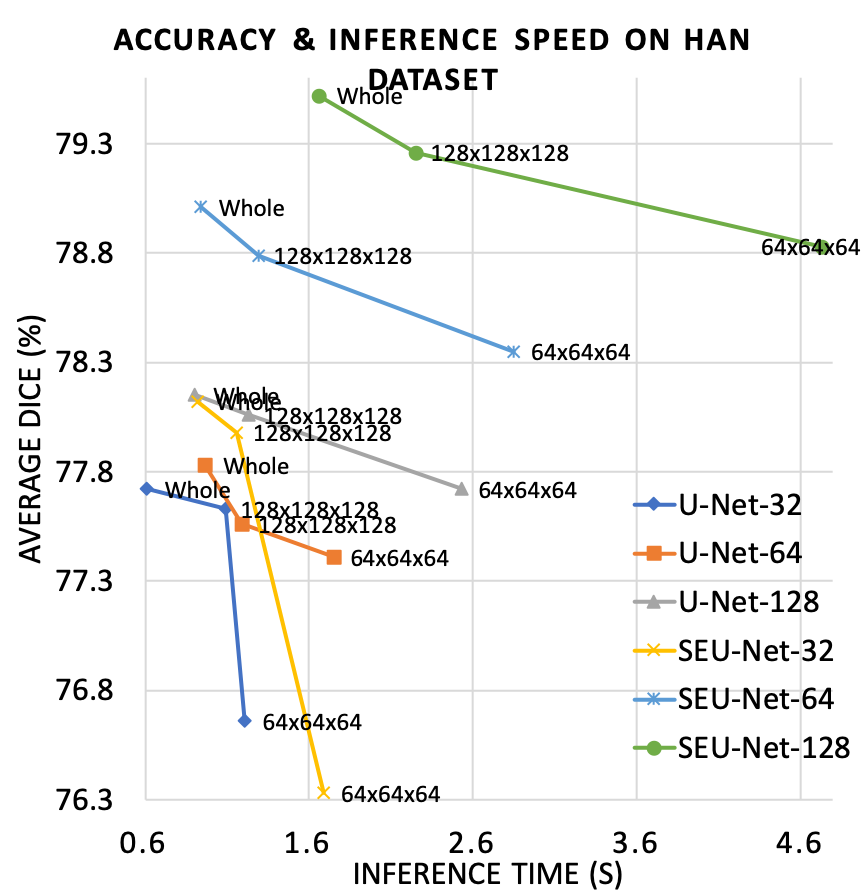}
		 		\end{minipage}
	 		\begin{minipage}{0.5\linewidth}
	 			\includegraphics[width=\textwidth,trim=2 2 2 2,clip]{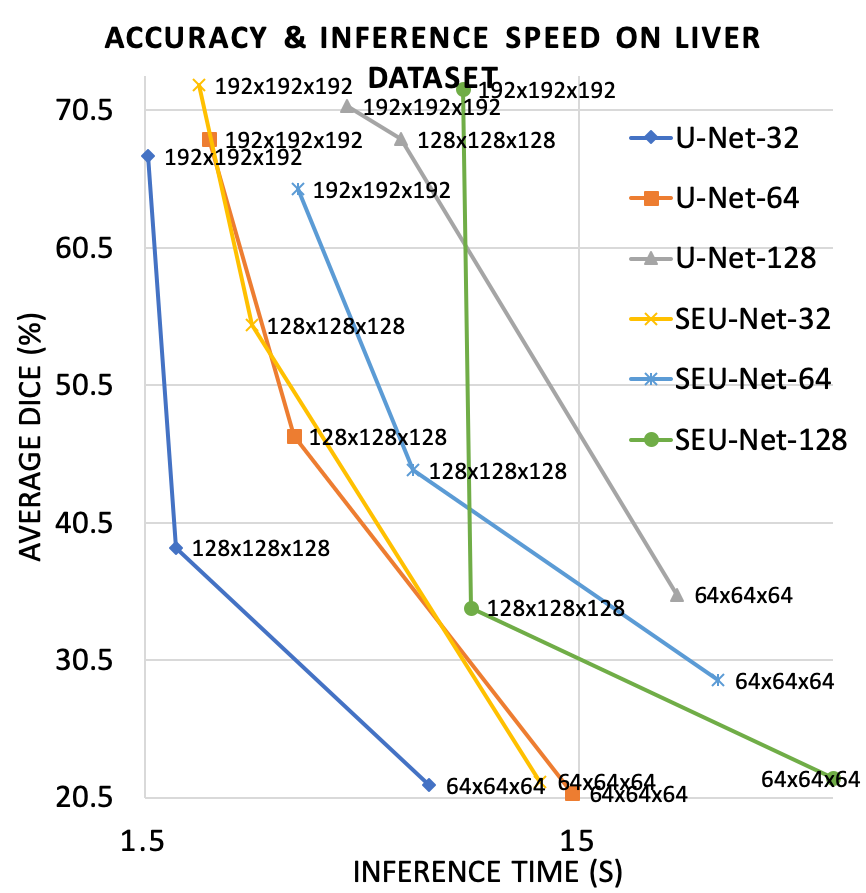}
	 		\end{minipage}

\caption{Segmentation accuracy (Dice coefficient, \%) and inference time (s) comparisons among 3D U-Net and 3D SEU-Net of different sizes (\#filters in the first convolutional layer: 32, 64, 128) and different input sizes (64$\times$64$\times$64, 128$\times$128$\times$128, whole image or 192$\times$192$\times$192) on HaN nine organ auto-segmentation and decathlon liver and tumor segmentation datasets. Large model and input yield better segmentation accuracy consistently, and large input significantly decreases inference time.} \label{fig:acc_speed}
\end{figure}

In this work, we investigate the impact of model size and input size in medical image analysis. We choose 3D U-Net~\cite{ronneberger2015u} and the other advanced U-Net, 3D Squeeze-and-Excitation U-Net (SEU-Net)~\cite{hu2018squeeze} in AnatomyNet~\cite{zhu2019anatomynet}, and validate them on large image segmentation tasks, i.e., head and neck (HaN) multi-organ segmentation~\cite{zhu2019anatomynet} and decathlon liver and tumor segmentation~\cite{simpson2019large}. Considering the flexibility and efficiency, we design a parallel U-Net based on GPipe~\cite{huang2019gpipe} as the back-end parallelism. In the training, we employ existing well-designed adaptive weighted loss in~\cite{zhu2019anatomynet} and design a curriculum training strategy based on different input sizes. Specifically, we sequentially fit the model with small patches for training in the first stage, medium patches thereafter, and large input lastly.  We conduct extensive experiments, and conclude that, employing large models and input context increases segmentation accuracy. Large input also reduces inference time significantly by leveraging automated model parallelism in Fig.~\ref{fig:acc_speed}.


%
%

\section{Method}
Considering flexibility and efficiency, we employ GPipe~\cite{huang2019gpipe} as the backend parallelism. The model parallelism is introduced in Section~\ref{sec:mp}. We describe how to design a parallel U-Net in Section~\ref{sec:p_u_net}. How to train the large models with large context input is introduced in Section~\ref{sec:train}.
\subsection{Automated Model Parallelism}\label{sec:mp}
\begin{figure}[t]
\begin{center}
\includegraphics[width=0.7\textwidth,trim=4 14 2 2,clip]{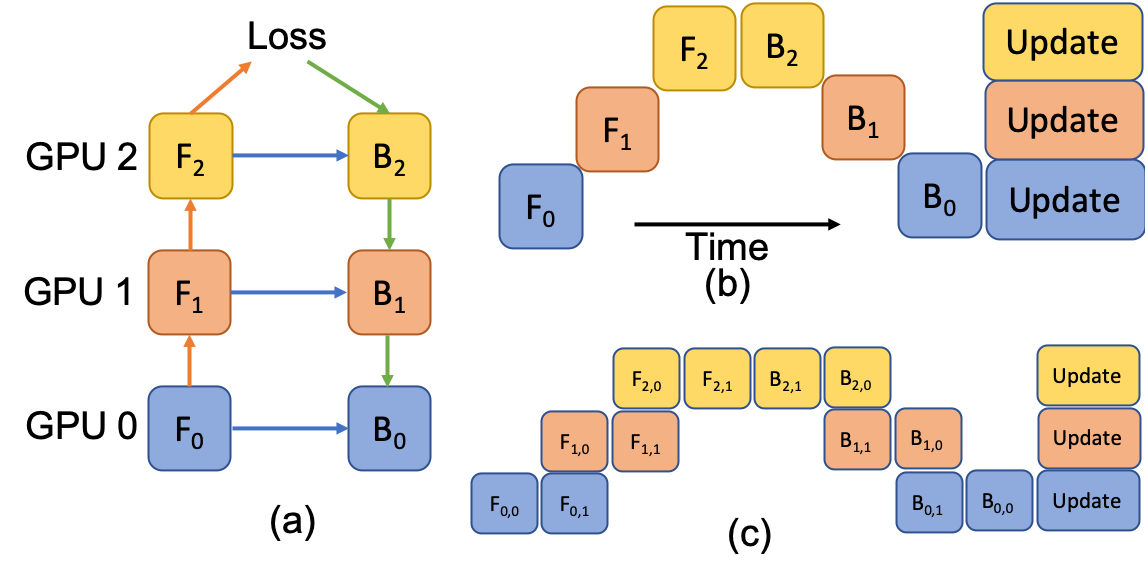}
\end{center}
\caption{(a) A deep model is partitioned across three GPUs. $F_k$ is the forward function of the $k$-th cell. $B_k$ is the back-propagation function which relies on both $B_{k+1}$ from upper layer and feature $F_k$. (b) Conventional model parallelism has low device utilization because of dependency of the model. (c) Pipeline parallelism splits the input mini-batch to smaller micro-batches (two micro-batches in the figure) and enables different devices to run micro-batches simultaneously. Synchronized gradient calculation can be applied lastly.} \label{fig:framework}
\end{figure}
Deep networks can be defined as a sequential model of $L$ layers. Each layer $L_i$ can be modeled by a forward computation function $f_i$ with parameters $w_i$. Given the number of partitions $K$, i.e., the number of GPUs typically, the model can be partitioned into $K$ parts as illustrated in~Fig. 2 (a). Specifically, let part $p_k$ consist of consecutive layers from layer $L_i$ to layer $L_j$. The parameters of part $p_k$ is the union of parameters $w_i, w_{i+1}, \dots, w_j$, and the forward function can be derived sequentially
\begin{equation}
F_k = f_j \circ f_{j-1} \circ \dots \circ f_i.
\end{equation}
According to the chain rule in the gradient calculation, the back-propagation function $B_k$ can be derived from $F_k$ by automated symbolic differentiation in the existing deep learning packages, e.g., PyTorch. 

In the forward pass, GPipe~\cite{huang2019gpipe,torchgpipe} first splits the input mini-batch of size $N$ to $M$ micro-batches as illustrated in Fig~\ref{fig:framework} (c). Micro-batches are pipelined through $K$ devices by model parallelism sequentially as illustrated in Fig~\ref{fig:framework} (b). This micro-batch splitting in Fig~\ref{fig:framework} (c) has a higher device utilization than conventional model parallelism in Fig~\ref{fig:framework} (b). After forward pass of all the micro-batches in the current mini-batch, gradients from all $M$ micro-batches are accumulated synchronously and back-propagation is applied to update model parameters. GPipe reduces space complexity from $O(N \times L)$ to $O(N + \frac{L}{K} \times \frac{N}{M})$, where $\frac{L}{K}$ is the size of layers per partition and $\frac{N}{M}$ is the micro-batch size~\cite{huang2019gpipe}.

\subsection{Parallel U-Net}\label{sec:p_u_net}
The pipeline parallelism is extremely simple and intuitive, and it is flexible and can be easily used to design various parallel algorithms. To use GPipe, we only need to 1) set the number of partitions $K$, which is the number of GPUs typically, 2) set the number of micro-batches $M$, which can also be set as the number of GPUs for efficiency, 3) modify the network into sequential layers. Next, we describe how to design a parallel U-Net.

We employ the conventional U-Net~\cite{ronneberger2015u}, which can be divided into three parts: an encoder $E$ with five blocks $e_1, e_2, \dots, e_5$ from input sequentially, a decoder $D$ with four blocks $d_5, d_4, \dots, d_1$, and four skip connections $s_1, s_2, \dots, s_4$. The U-Net can be formulated
\begin{equation}
\begin{aligned}
E &= e_5 \circ e_4 \circ \dots \circ e_1, \quad d_i = s_i(e_i, d_{i+1}), i = 1, 2, \dots, 4, \\
D &= d_1 \circ d_2 \circ \dots \circ d_5,
\end{aligned}
\end{equation}
where $s_i$ is typically a concatenation along channel dimension. The input of encoder $E$ is the image, and the input of decoder block $d_5$ is the output of encoder. We can then add a softmax function after decoder $D$ for segmentation.

The main challenge of pipeline-based parallel U-Net is the dependency of intermediate encoder in the skip connection $s_i$. GPipe requires that the model needs to be implemented in a sequential way. However, each $e_i, i = 1, 2, \dots, 4,$ is used in both encoder and decoder, which affects automated partition in GPipe. We can remove the dependency and modify U-Net by duplicating the output of each encoder $e_i = \{e_{i,0}, e_{i,1}\}, i = 1, 2, \dots, 4$. Specifically, the sequential U-Net can be derived
\begin{equation}
\begin{aligned}
E &= e_{5} \circ e_{4,0} \dots \circ e_{2,0} \circ e_{1,0}, \quad d_i = s_i(e_{i,1}, d_{i+1}), i = 1, 2, \dots, 4, \\
D &= d_1 \circ d_2 \circ \dots \circ d_5.
\end{aligned}
\end{equation}

The temporary variable $e_{i,1}$ breaks the dependency in the skip connection and facilitates the automated partition in automated parallelism of GPipe. We can employ the existing GPipe algorithm to implement parallel U-Net based on the designed sequential U-Net.

\subsection{Learning Large Models}\label{sec:train}
Leveraging the powerful tool of parallel U-Net, we investigate the impact of model size and input context size. Although previous study demonstrates large input size increases segmentation accuracy because of large context~\cite{isensee2019nnu}, it also decreases the variation of training input and aggravates the extremely imbalance issue between background and the small subjects. From model size's perspective, large model consists of more parameters which typically require more various data to fit. Therefore, designing a learning strategy is essential to fully exploit the power of large input with more context information.

Inspired by the learning theory, i.e. curriculum learning~\cite{bengio2009curriculum}, we can fit easy data/task into the network first and let the network to solve hard task later. Learning from smaller patches is easier, because smaller patches can be sampled with less imbalance and the lower dimension of smaller patches consists of less structures to learn for structured tasks, e.g., image segmentation. In practice, we firstly sample small positive patches (size of 64$\times$64$\times$64) to train the model in the initial stage. In the second stage, we sample medium positive patches (size of 128$\times$128$\times$128) to train the model. Finally, we use the largest patch to train the model. In this way, we can fully train models with large input patches in a practical way.
\section{Experiments}
\begin{table}[t]
\caption{Dice coefficient (\%) achieved on the HaN test set using different sizes of U-Nets and inputs.}\label{tab:unet_han}
\begin{tabular}{|l|l|l|l|l|l|l|l|l|l|l|}
\hline
Models & BS &  CH &  MA &  OL &  OR & PL & PR & SL &  SR & Average $\uparrow$\\
\hline
U-Net-32 ($64^3$) & 84.23 & 48.87 & 89.75 & 69.11 & 68.28 & 87.43 & 85.48 & 79.36 & 77.41 & 76.66\\
U-Net-64 ($64^3$) & 84.28 & 46.21 & 91.55 & 70.34 & 69.92 & 87.76 & 85.98 & 81.46 & 79.23 & 77.41\\
U-Net-128 ($64^3$) & 84.58 & 48.52 & 91.12 & 71.04 & 69.28 & 87.76 & 85.78 & 81.34 & 80.03 & 77.72 \\ \hline

U-Net-32 ($128^3$) & 84.23 & 53.30 & 91.97 & 70.29 & 68.40 & 87.43 & 85.48 & 79.36 & 78.17 & 77.63 \\
U-Net-64 ($128^3$) & 84.71 & 46.21 & 92.47 & 70.34 & 69.92 & 87.76 & 85.98 & 81.46 & 79.23 & 77.56\\
U-Net-128 ($128^3$) & 84.84 & 48.52 & 93.71 & 71.04 & 69.28 & 87.76 & 85.78 & 81.57 & 80.03 & 78.06\\ \hline %

U-Net-32 (Whole) &84.23&53.30&91.97&70.29&68.40&87.43&85.48&79.36&79.02&77.72\\
U-Net-64 (Whole) &84.71&48.59&92.47&70.34&69.92&87.76&85.98&81.46&79.23&77.83\\
U-Net-128 (Whole) &84.84 & 48.52 & 93.71 & 71.04 & 70.09 & 87.76 & 85.78 & 81.57 & 80.03 & 78.15\\ \hline %
\end{tabular}
\end{table}

\begin{table}[t]
\caption{Dice coefficient (\%) achieved on the HaN test set using different sizes of SEU-Nets and inputs.}\label{tab:resunet_han}
\begin{tabular}{|l|l|l|l|l|l|l|l|l|l|l|}
\hline
Models & BS &  CH &  MA &  OL &  OR & PL & PR & SL &  SR & Average $\uparrow$\\
\hline
AnatomyNet~\cite{zhu2019anatomynet} & 86.65 & 53.22 & 92.51 & 72.10 & 70.64 & 88.07 & 	87.35 & 81.37 & 81.30 & 79.25 \\ \hline
SEU-Net-32 ($64^3$) & 84.07 & 47.09 & 90.12 & 68.58 & 69.73 & 87.14 & 85.21 & 79.20 & 75.81 & 76.33\\
SEU-Net-64 ($64^3$) & 85.49 & 50.32 & 92.45 & 71.93 & 69.94 & 88.24 & 86.27 & 81.15 & 79.37 & 78.35\\
SEU-Net-128 ($64^3$) & 86.38 & 51.85 & 93.55 & 70.62 & 70.08 & 88.11 & 85.99 & 81.79 & 81.13 & 78.83\\ \hline

SEU-Net-32 ($128^3$) & 85.76 & 50.52 & 92.91 & 70.76 & 69.73 & 87.31 & 85.86 & 81.03 & 77.95 & 77.98\\
SEU-Net-64 ($128^3$) &85.73&50.37&94.26&71.97&71.09&88.34&86.58&81.15&79.64&78.79\\
SEU-Net-128 ($128^3$) &86.38&51.85&93.87&71.63&70.44&88.11&86.75&81.79&82.48&79.26\\ \hline

SEU-Net-32 (Whole) & 85.76&51.27&92.91&70.76&69.73&87.31&85.86&81.03&78.43&78.12\\
SEU-Net-64 (Whole) &85.73 & 52.29 & 94.26 & 71.97 & 71.09 & 88.34 & 86.58 & 81.15 & 79.64 & 79.01\\
SEU-Net-128 (Whole) &86.38 & 51.85 & 93.87 & 73.70 & 70.44 & 88.26 & 86.75 & 81.96 & 82.48 & \textbf{79.52}\\ \hline 
\end{tabular}
\end{table} 
\begin{table}[t]
\caption{Average inference time (s) per test image achieved on the HaN test set using different sizes of networks and inputs.}\label{tab:speed_han}
\begin{tabular}{|l|l||l|l|}
\hline
Models & Inference time $\downarrow$ & Models & Inference time $\downarrow$\\
\hline
U-Net-32 ($64^3$) 2$\times$16G& 1.21$\pm$0.07 & SEU-Net-32 ($64^3$) 2$\times$16G&1.69$\pm$0.17 \\
U-Net-64 ($64^3$) 4$\times$16G& 1.75$\pm$0.08 & SEU-Net-64 ($64^3$) 2$\times$32G& 2.85$\pm$0.13\\
U-Net-128 ($64^3$) 2$\times$32G& 2.53$\pm$0.04 & SEU-Net-128 ($64^3$) 4$\times$32G& 4.73$\pm$0.69\\ \hline

U-Net-32 ($128^3$) & 1.09$\pm$0.28  & SEU-Net-32 ($128^3$) &1.16$\pm$0.36 \\
U-Net-64 ($128^3$) & 1.19$\pm$0.16 & SEU-Net-64 ($128^3$) & 1.29$\pm$0.18\\
U-Net-128 ($128^3$) & 1.23$\pm$0.16 & SEU-Net-128 ($128^3$)& 2.25$\pm$0.13\\ \hline

U-Net-32 (Whole) & 0.61$\pm$0.07 & SEU-Net-32 (Whole) &0.92$\pm$0.07 \\
U-Net-64 (Whole) &  0.96$\pm$0.22 & SEU-Net-64 (Whole) &0.94$\pm$0.07\\
U-Net-128 (Whole) &  0.90$\pm$0.14 & SEU-Net-128 (Whole)& 1.66$\pm$0.14\\ \hline
\end{tabular}
\end{table} 
We use two datasets to investigate the impact of large models and large input context for segmentation, the head and neck (HaN)  and decathlon liver datasets. The HaN dataset consists of whole-volume computed tomography (CT) images with manually generated binary masks of nine anatomies, i.e., brain stem~(BS), chiasm~(CH), mandible~(MD), optic nerve left~(OL), optic nerve right~(OR), parotid gland left~(PL), parotid gland right~(PR), submandibular gland left~(SL), and submandibular gland right~(SR). We download the publicly available preprocessed data from AnatomyNet~\cite{zhu2019anatomynet}, which includes three public datasets: 1) MICCAI Head and Neck Auto Segmentation Challenge 2015~\cite{raudaschl2017evaluation}; 2) the Head-Neck Cetuximab collection from The Cancer Imaging Archive (TCIA)~\cite{clark2013cancer}; 3) the CT images from four different institutions in Qu\'ebec, Canada~\cite{vallieres2017radiomics}, also from TCIA. We use the dataset directly for fair comparison with benchmark methods. The dataset consists of 261 training images with missing annotations and ten test samples consisting of all annotations of nine organs. The largest image size can be 352$\times$256$\times$288. We use the same data augmentation techniques in~\cite{zhu2019anatomynet}. 

The other dataset is 3D liver and tumor segmentation CT dataset from the medical segmentation decathlon~\cite{simpson2019large}. We randomly split the dataset into 104 training images and 27 test images. We re-sample the CT images to 1$\times$1$\times$1 $\mathrm{mm}^3$ spacing. To focus on the liver region, we clip the voxel value within range $[-21, 89]$ and linearly transform each 3D image into range $[0, 1]$. In the training, we randomly flip and rotation 90 degrees in XY space with probability 0.1. We further add uniform random noise $[-0.2, 0.2]$ to augment the training data. The largest image size can be 512$\times$512$\times$704. We will release the script and data splitting for reproducibility. 


In the training, for the largest input, we use batch size of one and RMSProp optimizer~\cite{tieleman2012lecture} with 300 epochs and learning rate of 1$\times$$10^{-3}$. For training with patch size 128$\times$128$\times$128, we use batch size of four and 1200 epochs. For training with patch size 64$\times$64$\times$64, we use batch size of 16 and 4800 epochs. For U-Net-32 and Squeeze-and-Excitation U-Net (SEU-Net-32), the number of filters in each convolution of the first encoder block is 32. We increase the number of filters to 64 and 128 to investigate the impact of increasing model size. In the encoder of each model, the number of filters are doubled with the increase of encoder blocks accordingly. The decoder is symmetric with the encoder.

We employ two networks, 3D U-Net and 3D SEU-Net, to investigate the impact of model size and input context size in table~\ref{tab:unet_han} and~\ref{tab:resunet_han} on HaN dataset. With the increase of model size and input size, the segmentation accuracy increases consistently for both U-Net and SEU-Net. The SEU-Net-128 with whole image as input achieves better performance than AnatomyNet searching different network structures~\cite{zhu2019anatomynet}. The reason for the accuracy improvement is that large input and model yield big context and learning capacity, respectively. We investigate the impact of large input on inference time by averaging three rounds of inferences in table~\ref{tab:speed_han}. Using large input in the inference reduces the inference time significantly because it reduces the number of inference rounds. Results on liver and tumor segmentation task validate large input increases segmentation accuracy and reduces the inference time in table~\ref{tab:unet_liver} and~\ref{tab:speed_liver}.

\begin{table}[t]
\caption{Dice coefficientt (\%) achieved on the Decathlon liver segmentation test set using different sizes of inputs and U-Nets and SEU-Nets.}\label{tab:unet_liver}
\begin{tabular}{|l|l|l|l||l|l|l|l|}
\hline
Models & Liver & Tumor & Average $\uparrow$ & Models & Liver & Tumor & Aevage $\uparrow$\\
\hline
U-Net-32 ($64^3$) & 4.76 & 38.06 & 21.41 & SEU-Net-32 ($64^3$) & 0.73 & 42.56 & 21.65\\
U-Net-64 ($64^3$) &9.70 & 31.96 & 20.83 & SEU-Net-64 ($64^3$) &11.90 & 46.19 & 29.05\\
U-Net-128 ($64^3$) &34.52&35.99&35.26 & SEU-Net-128 ($64^3$) & 0.34  & 43.44 & 21.89\\ \hline %

U-Net-32 ($128^3$) & 26.23 & 51.12 & 38.68 & SEU-Net-32 ($128^3$) & 58.88 & 50.83 & 54.86 \\
U-Net-64 ($128^3$) & 40.95 & 52.63 & 46.79 & SEU-Net-64 ($128^3$) &  38.38 & 50.25 & 44.32 \\
U-Net-128 ($128^3$) & 84.83 & 51.98 & 68.41 & SEU-Net-128 ($128^3$) & 20.20  & 48.44 & 34.32\\ \hline 

U-Net-32 ($192^3$) & 82.83 & 51.57 & 67.20 & SEU-Net-32 ($192^3$) &89.25&55.38&72.32\\
U-Net-64 ($192^3$) & 91.58 & 45.29 & 68.44 & SEU-Net-64 ($192^3$) &  77.66 & 51.93 & 64.80\\ 
U-Net-128 ($192^3$) & 90.99 & 50.67 & 70.83 & SEU-Net-128 ($192^3$) &  87.61 & 56.48 & 72.05\\ \hline 
\end{tabular}
\end{table}

%
%

\begin{table}[t]
\caption{Average inference time (s) per test image achieved on the Decathlon liver segmentation test set using different sizes of networks and inputs.}\label{tab:speed_liver}
\begin{tabular}{|l|l||l|l|}
\hline
Models & Inference time $\downarrow$ & Models & Inference time $\downarrow$\\
\hline
U-Net-32 ($64^3$) 2$\times$16G & 6.78$\pm$0.06 & SEU-Net-32 ($64^3$) 4$\times$16G& 12.23$\pm$0.08 \\ 
U-Net-64 ($64^3$) 4$\times$16G& 14.52$\pm$0.02 & SEU-Net-64 ($64^3$) 2$\times$32G& 31.47$\pm$0.16 \\
U-Net-128 ($64^3$) 4$\times$32G& 25.37$\pm$1.10 & SEU-Net-128 ($64^3$) 8$\times$32G& 57.99$\pm$11.08 \\ \hline

U-Net-32 ($128^3$) & 1.77$\pm$0.42 & SEU-Net-32 ($128^3$) & 2.64$\pm$0.06 \\ 
U-Net-64 ($128^3$) & 3.30$\pm$0.52 & SEU-Net-64 ($128^3$) & 6.23$\pm$0.17 \\
U-Net-128 ($128^3$) & 5.84$\pm$0.21 & SEU-Net-128 ($128^3$)& 8.49$\pm$0.08\\ \hline

U-Net-32 ($256^3$) & 1.52$\pm$0.58 & SEU-Net-32 ($256^3$) & 2.00$\pm$0.20 \\ 
U-Net-64 ($256^3$) & 2.11$\pm$0.10  & SEU-Net-64 ($256^3$) &  3.37$\pm$0.10\\
U-Net-128 ($256^3$) &4.39$\pm$0.25 & SEU-Net-128 ($256^3$)& 8.10$\pm$0.50\\ \hline

\end{tabular}
\end{table}


%
\section{Conclusion}
In this work, we try to investigate the impact of model size and input context size on two medical image segmentation tasks. To run large models and large input in the GPUs, we design a parallel U-Net with sequential modification based on an automated parallelism. Extensive results demonstrate that, 1) large model and input increases segmentation accuracy, 2) large input reduces inference time significantly. The Large deep networks with Automated Model Parallelism (LAMP) can be a useful tool for many medical image analysis tasks such as large image registration~\cite{zhu2020neurreg,zhu2019neural}, detection~\cite{zhu2018deeplung,zhu2018deepem} and neural architecture search.  

\bibliographystyle{splncs04}
\bibliography{mybibliography}
%





\appendix
\section{Appendix: Design of LAMP}
\begin{figure}
\begin{center}
\includegraphics[width=0.8\textwidth,trim=2 2 2 2,clip]{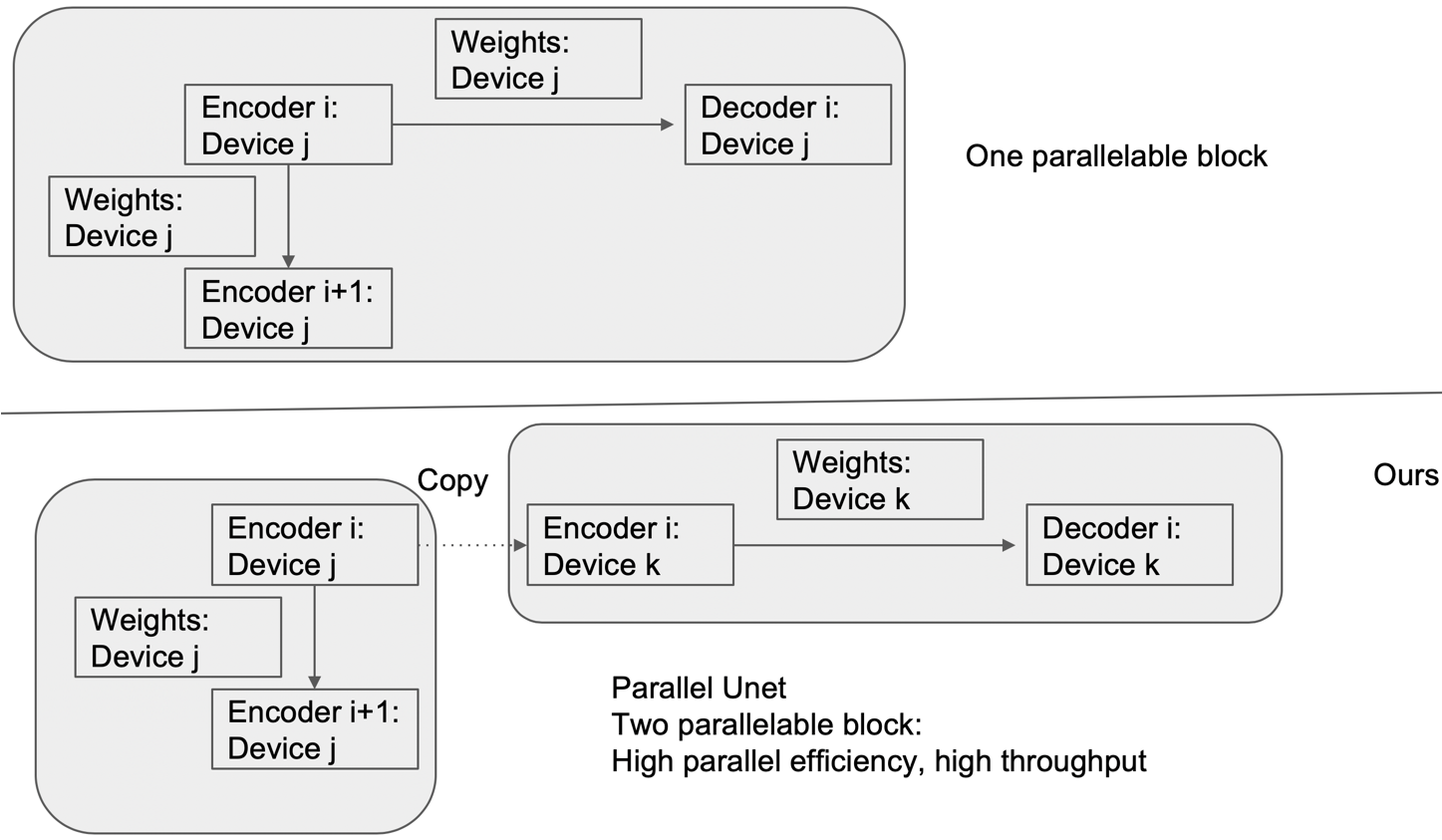}
\end{center}
\caption{Up: The long range skip-connection hinders the parallelism in the U-Net. Bottom: We explicitly construct a variant of U-Net to remove the long range dependency in the U-Net. The parallel U-Net has higher parallel efficiency.} \label{fig:design}
\end{figure}

The figure~\ref{fig:design} shows we reduce the dependency of long range skip-connection (Up) by separating it to two blocks (Bottom). Through the design of LAMP, the parallel U-Net achieves more parallel blocks, which lead to high throughput. We proof this in the next section.
\section{Appendix: Proof for High Throughput of LAMP}
\begin{figure}
\begin{center}
\includegraphics[width=0.8\textwidth,trim=2 2 2 2,clip]{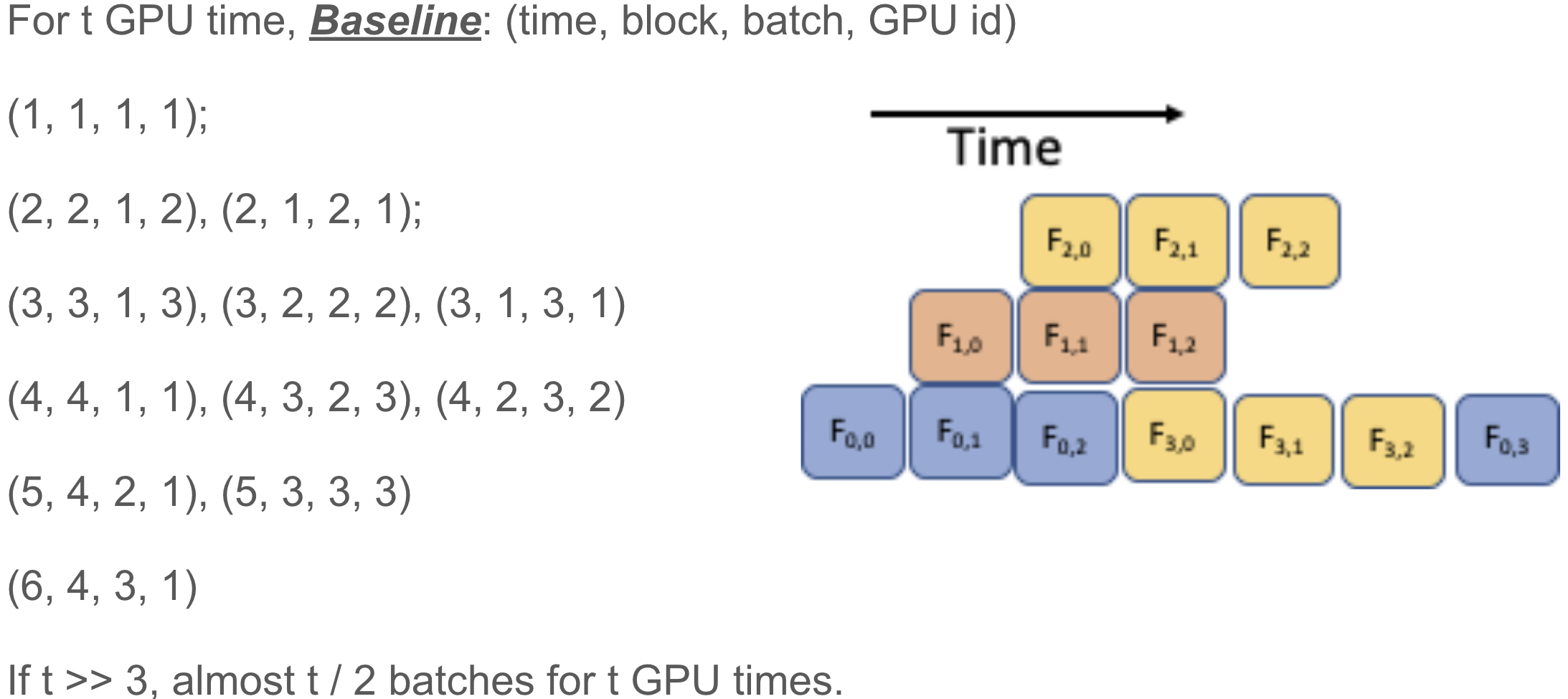}
\end{center}
\caption{In the conventional U-Net based on pipeline parallelism, using three devices processes $t/2$ batches in $t$ device time.} \label{fig:proof}
\end{figure}

\begin{figure}
\begin{center}
\includegraphics[width=0.8\textwidth,trim=2 2 2 2,clip]{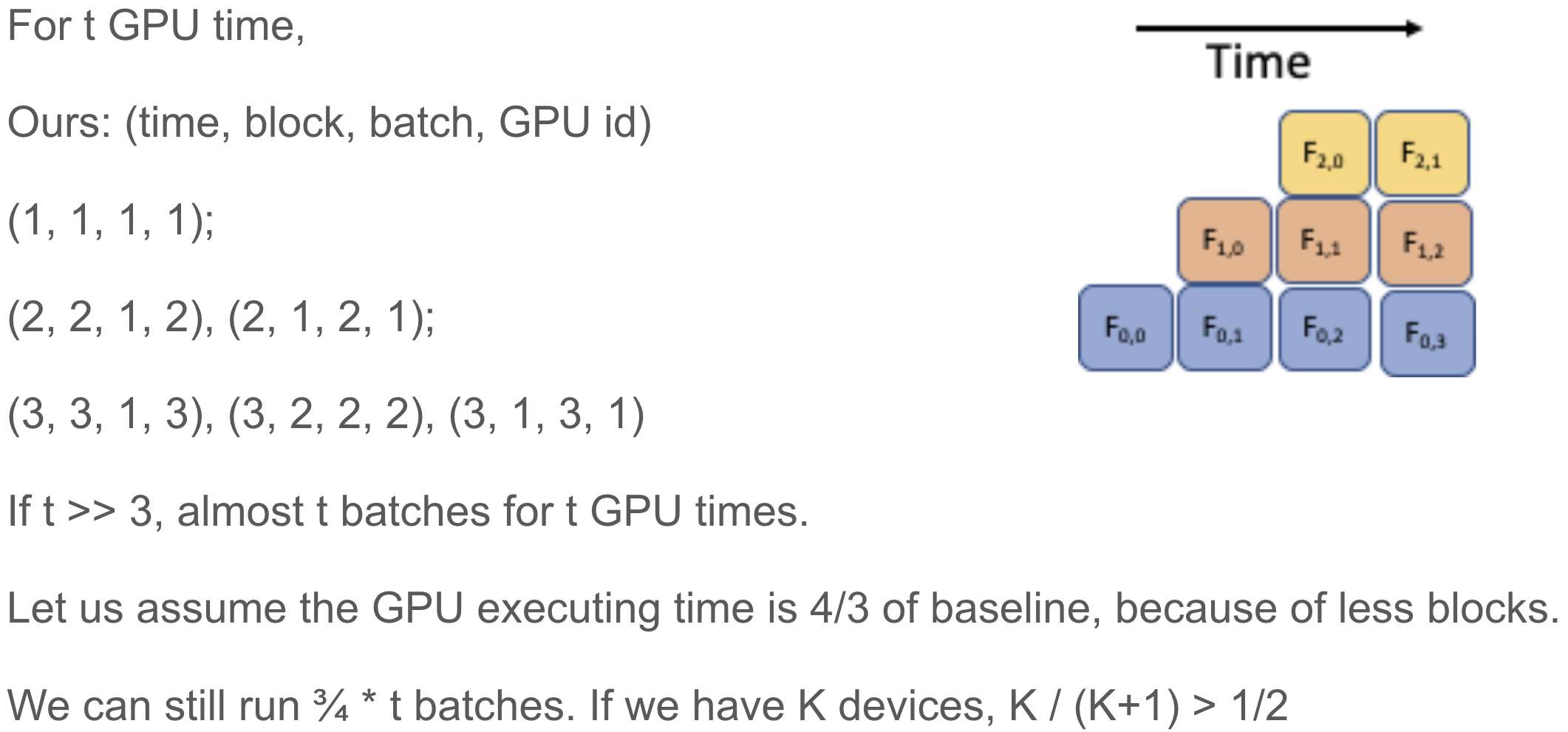}
\end{center}
\caption{In the parallel U-Net based on LAMP parallelism, using three devices processes $3t/4$ batches in $t$ device time. Parallel U-Net based on LAMP has a higher throughput.} \label{fig:proof_lamp}
\end{figure}

We demonstrate the parallel U-Net with LAMP has higher throughput in Fig.~\ref{fig:proof} and Fig.~\ref{fig:proof_lamp}.

\end{document}